\documentclass[lettersize,journal]{IEEEtran}
\usepackage{algorithm,algorithmic,amsbsy,amsmath,amssymb,epsfig,bbm,mathrsfs,fancyhdr,fancyvrb,subfigure,url,cite,multirow,xcolor}

\usepackage[hidelinks]{hyperref}
\usepackage{graphicx}
\usepackage{epstopdf}
\usepackage{setspace}
\usepackage{subfigure}
\usepackage{caption}
\usepackage{makecell}
\usepackage{booktabs}
\usepackage[table]{xcolor}
\allowdisplaybreaks
\usepackage{svg}

\newcommand{\beq}{\begin{equation}}
\newcommand{\eeq}{\end{equation}}
\newcommand{\beqn}{\begin{eqnarray}}
\newcommand{\eeqn}{\end{eqnarray}}
\usepackage{graphicx}
\usepackage{amsmath, amssymb}

\providecommand{\theoremname}{\textbf{Theorem}}
\providecommand{\propositionname}{\textbf{Proposition}}
\providecommand{\remarkname}{\textbf{Remark}}
\providecommand{\lemmaname}{\textbf{Lemma}}
\providecommand{\corollaryname}{\textbf{Corollary}}
\providecommand{\Definition}{\textbf{Definition}}
\usepackage{booktabs}
\usepackage{tabularx} 
\newcolumntype{Y}{>{\raggedright\arraybackslash}X}

\begin{document}

\title{Storage-Scalable Progressive Semantic Communication via Knowledge-Base Reuse}

\author{Heng Zhu, Ye Liu, Kun Zhu,~\IEEEmembership{Member,~IEEE}, Feifei Song

\thanks{Heng Zhu, Ye Liu, Kun Zhu and Feifei Song are with the College of Computer Science and Technology, Nanjing University of Aeronautics and Astronautics, Nanjing 210016, China (e-mail: \{zhuheng, sz2516087, zhukun, songff\}@nuaa.edu.cn).}
}

\maketitle

\begin{abstract}
Existing knowledge-base-assisted semantic communication schemes commonly
adopt either single knowledge-base quantization (SKBQ) or
multi-knowledge-base residual quantization (MKBQ). SKBQ incurs limited
storage overhead but has restricted quantization capacity, whereas MKBQ
supports progressive refinement by assigning an independent knowledge base
(KB) to each stage, causing the KB storage to grow linearly with the
transmission depth. To address this problem, we propose storage-scalable
knowledge-base reuse quantization (SSKBQ), which reuses a compact set of KBs across multiple residual refinement stages and thereby decouples the number of transmission stages from the number of maintained KBs. A stage-aware residual supervision mechanism is further introduced to regularize
intermediate quantized representations and encourage progressive
refinement. Experimental results demonstrate that KB reuse
provides an effective solution to the storage scalability problem while maintaining competitive progressive reconstruction performance.
\end{abstract}


\section{Introduction}
\IEEEPARstart{S}{emantic} communication has recently emerged as a promising paradigm and has attracted extensive research interest \cite{shao2024theory, luo2022semantic, yang2022semantic}. Unlike conventional communication systems that pursue bit-level fidelity, semantic communication focuses on delivering task-relevant information. By removing task-irrelevant redundancy, it can significantly reduce transmission rates while preserving task performance.

Despite these advantages, early studies reveal a fundamental challenge: without carefully designed encoding and decoding mechanisms, semantic communication may even require higher transmission rates than conventional schemes. This issue largely stems from the use of deep neural networks for semantic feature extraction, where integer-valued inputs are transformed into high-dimensional floating-point representations. According to the IEEE 754 standard \cite{markstein2008new}, a double-precision floating-point number typically occupies 64 bits, whereas an integer often requires only 8 bits. Consequently, directly transmitting semantic features requires substantial compression to maintain the same transmission cost, which is often impractical.

To alleviate this issue, quantization has been introduced into semantic encoding. A representative approach is SKBQ \cite{van2017neural}, where semantic features are mapped to the nearest codewords in a predefined KB. Instead of transmitting floating-point features, the corresponding integer indices are conveyed, thereby reducing transmission rates. However, the use of a single KB limits the representation capacity and may
lead to considerable quantization distortion. Moreover, it typically supports only single-stage transmission and does not naturally support progressive refinement.

To enable progressive refinement, MKBQ schemes have been proposed \cite{adiban2025s}. These methods progressively quantize residual errors across multiple KBs, enabling progressive reconstruction through multi-stage transmission. However, the storage overhead grows linearly with the number of stages, since each stage requires an independent KB. When each KB is large, the cumulative storage cost becomes prohibitive.

To resolve this trade-off, we propose an SSKBQ scheme for progressive
semantic communication. The key idea is to decouple transmission stages
from dedicated KBs by reusing a compact set of KBs across multiple residual refinement stages, thereby enabling progressive reconstruction without
linear KB storage growth. \textcolor{black}{In addition, a stage-aware
residual supervision mechanism is designed to regularize intermediate
quantized representations and encourage progressive refinement across
stages.} Experimental results show that the proposed scheme provides a
favorable trade-off between progressive reconstruction performance and KB
storage overhead compared with SKBQ and MKBQ.

\begin{figure}[!t]
    \centering
    \scriptsize
    \setlength{\belowcaptionskip}{-10pt}
    \includegraphics[width=3.5in]{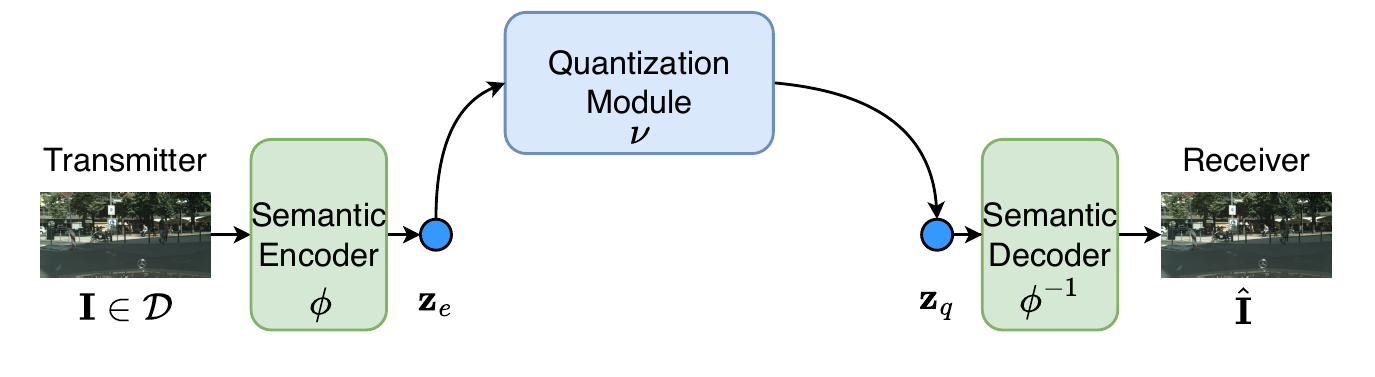}  
    \caption{\textcolor{black}{End-to-End semantic communication system for image reconstruction.}}
    \label{net_framework}
\end{figure}

\section{Semantic Communication System}

In this work, we consider an end-to-end (E2E) semantic communication system for image reconstruction, as illustrated in Fig.~\ref{net_framework}. \textcolor{black}{We focus on the semantic-layer design of storage-scalable progressive transmission rather than physical-layer transmission or channel-robust modulation and coding.} Let $\mathbf{I} \in \mathbb{R}^{3 \times H \times W}$ denote an input image sampled from dataset $\mathcal{D}$, where the three dimensions represent the RGB channels, height, and width, respectively.

At the transmitter, the image is processed by a semantic encoder $\phi(\cdot)$ to extract task-relevant semantic features:
\begin{equation}
\mathbf{z}_e = \phi(\mathbf{I}), \quad
\mathbf{z}_e \in \mathbb{R}^{c \times h \times w},
\end{equation}
where $c$, $h$, and $w$ denote the number of channels, height, and width of the feature map, respectively. The extracted semantic feature $\mathbf{z}_e$ is then transmitted to the receiver. However, direct transmission of $\mathbf{z}_e$ in floating-point format incurs substantial transmission overhead. \textcolor{black}{To reduce this overhead, a quantization module $\nu(\cdot)$ is introduced to quantize $\mathbf{z}_e$ as}
\begin{equation}
\mathbf{z}_q = \nu(\mathbf{z}_e), \quad
\mathbf{z}_q \in \mathbb{R}^{c \times h \times w},
\end{equation}
which is then transmitted to the receiver. \textcolor{black}{We assume perfect physical-layer transmission. Accordingly, channel coding, modulation, equalization, and error-control mechanisms are abstracted as a reliable delivery interface.} At the receiver, the received $\mathbf{z}_q$ is fed into the semantic decoder $\phi^{-1}(\cdot)$ to reconstruct the image:
\begin{equation}
\hat{\mathbf{I}} = \phi^{-1}(\mathbf{z}_q), \quad
\hat{\mathbf{I}} \in \mathbb{R}^{3 \times H \times W}.
\end{equation}

\section{Storage-Scalable Knowledge-Base Reuse Scheme}

\subsection{Knowledge-Base Reuse Scheme}

\begin{figure}[!t]
    \centering
    \scriptsize
    \setlength{\belowcaptionskip}{-10pt}
    \includegraphics[width=2.8in]{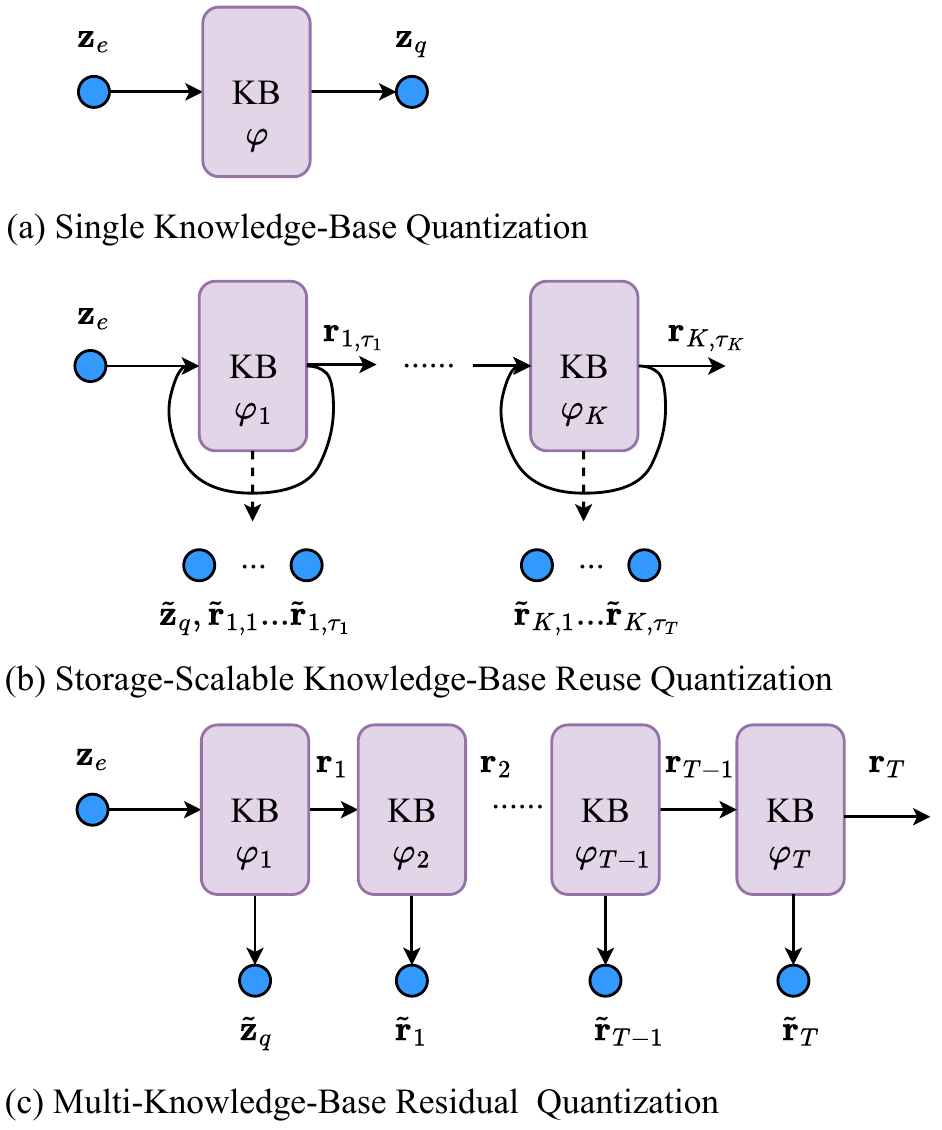}
    \caption{\textcolor{black}{Knowledge-base quantization schemes.}}
    \label{KBs}
\end{figure}

A straightforward implementation of the quantization module in
Fig.~\ref{net_framework} is based on vector quantization. \textcolor{black}{Specifically, the quantization module $\nu(\cdot)$ is parameterized by a learnable KB $\boldsymbol{\varphi}\in\mathbb{R}^{N\times c}$, which contains $N$ codewords of dimension $c$. The semantic feature $\mathbf{z}_e$ is reshaped into $hw$ vectors $\{\mathbf{z}_e^i\}_{i=1}^{hw}$, each of dimension $c$.} For each vector, the Euclidean distances to all codewords in the KB are computed, and the nearest codeword is selected as
\begin{equation}
\mathbf{z}_q^i=\boldsymbol{\varphi}^k,\quad
k=\arg\min_j
\left\|
\mathbf{z}_e^i-\boldsymbol{\varphi}^j
\right\|_2^2,
\end{equation}
where $\boldsymbol{\varphi}^j$ denotes the $j$-th codeword. By concatenating all $\mathbf{z}_q^i$ and reshaping them into size $c\times h\times w$, the quantized semantic feature $\mathbf{z}_q$ is obtained. As illustrated in Fig.~\ref{KBs}(a), SKBQ requires only one KB and thus incurs limited storage overhead. However, it provides limited reconstruction performance and does not support progressive transmission. Therefore, its reconstruction quality is often insufficient for high-resolution images.

To enable progressive transmission, MKBQ extends SKBQ by employing multiple homogeneous KBs. \textcolor{black}{As illustrated in Fig.~\ref{KBs}(c), the transmission process is divided into $T$ stages. Accordingly, the quantization module $\nu(\cdot)$ includes $T$ KBs, denoted by $\{\boldsymbol{\varphi}_j\}_{j=1}^{T}$, where $\boldsymbol{\varphi}_j\in\mathbb{R}^{N\times c}$.} The first KB generates an initial approximation $\tilde{\mathbf{z}}_q=\boldsymbol{\varphi}_1(\mathbf{z}_e)$, while the remaining KBs iteratively quantize the residuals. Specifically, the first residual is
$\mathbf{r}_1=\mathbf{z}_e-\tilde{\mathbf{z}}_q$, and each subsequent KB produces
$\tilde{\mathbf{r}}_i=\boldsymbol{\varphi}_{i+1}(\mathbf{r}_i)$, where
$\mathbf{r}_i=\mathbf{z}_e-\tilde{\mathbf{z}}_q-\sum_{j=1}^{i-1}\tilde{\mathbf{r}}_j$.
After $T$ stages, the final quantized semantic feature is
\begin{equation}
\mathbf{z}_q
=
\tilde{\mathbf{z}}_q
+
\sum_{i=1}^{T-1}\tilde{\mathbf{r}}_i.
\end{equation}

By adaptively selecting the number of participating KBs, MKBQ supports progressive transmission. However, two limitations remain. First, MKBQ does not guarantee monotonic refinement across transmission stages. Consequently, a later-stage reconstruction may not outperform an earlier-stage reconstruction, resulting in unstable progressive refinement. Second, the KB storage grows linearly with the number of transmission stages because each stage requires an independent KB containing numerous codewords. Since high-resolution image reconstruction generally requires more transmission stages, the resulting storage overhead can become prohibitively high.

To overcome these limitations, we propose the SSKBQ scheme, illustrated in Fig.~\ref{KBs}(b). Unlike MKBQ, which assigns an independent KB to each
transmission stage, SSKBQ reuses a compact set of KBs across multiple
residual refinement steps, thereby decoupling the number of maintained KBs
from the transmission depth.

Let $\tau_j$ denote the number of residual refinement steps assigned to the $j$-th KB. The first KB quantizes the semantic feature
$\mathbf{z}_e$ to obtain the initial approximation
$\tilde{\mathbf{z}}_q$ and is then reused to quantize the following
$\tau_1-1$ residuals. The second KB is subsequently reused to quantize the
residuals from step $\tau_1+1$ to $\tau_2$, and the remaining KBs follow the same strategy. Consequently, using only $K$ reusable KBs
($K\leq T$), the proposed scheme supports $T$ progressive transmission
stages through KB reuse. The resulting quantized semantic feature is
expressed as
\begin{equation}
\mathbf{z}_q
=
\tilde{\mathbf{z}}_q
+
\sum_{j=1}^{K}
\sum_{i=1}^{\tau_j}
\tilde{\mathbf{r}}_{j,i},
\end{equation}
where $\tilde{\mathbf{r}}_{j,i}$ denotes the quantization result produced
by the $j$-th KB at its $i$-th residual quantization step.

\subsection{Training Storage-Scalable Knowledge Base}
The semantic codec and the proposed storage-scalable KB are jointly optimized in an E2E manner. The overall objective consists of three components: the reconstruction loss $\mathcal{L}_{\mathrm{rec}}$, the quantization loss $\mathcal{L}_{\mathrm{vq}}$, and the residual supervision loss $\mathcal{L}_{\mathrm{res}}$. The reconstruction loss measures the distortion between the reconstructed image $\hat{\mathbf{I}}$ and the original image $\mathbf{I}$ using mean square error:
\begin{equation}
\mathbb{E}_{\mathbf{I} \sim \mathcal{D}}
\left[ \left\| \hat{\mathbf{I}} - \mathbf{I} \right\|_2^2 \right].
\end{equation}

The quantization loss enforces consistency between the semantic feature $\mathbf{z}_e$ and quantized feature $\mathbf{z}_q$:
\begin{equation}
\label{vq}
\mathrm{E}_{\mathbf{I}\sim\mathcal{D}} \bigg[\alpha\left\| \mathbf{z}_e - \mathrm{sg}[\mathbf{z}_q] \right\|_2^2 
+  \left\| \mathrm{sg}[\mathbf{z}_e] - \mathbf{z}_q \right\|_2^2\bigg],
\end{equation}
where $\mathrm{sg}(\cdot)$ denotes the stop-gradient operator. The first term updates the semantic encoder by encouraging $\mathbf{z}_e$ to approach $\mathbf{z}_q$, whereas the second term updates the KBs by aligning $\mathbf{z}_q$ with $\mathbf{z}_e$. The hyperparameter $\alpha$ balances these two effects. 

Although the quantization loss explicitly aligns the final quantized feature
$\mathbf z_q$ with the semantic feature $\mathbf z_e$, it imposes no
constraint on the intermediate quantized representations generated during
progressive transmission. Therefore, we introduce a residual supervision
loss to explicitly regularize the stage-wise refinement process. Let the intermediate quantized feature after $t$ refinement stages be defined as:
\begin{equation}
\mathbf{z}_q{(t)}
=
\tilde{\mathbf{z}}_q
+
\sum_{i=1}^{t}
\tilde{\mathbf{r}}_i.
\end{equation}
The residual supervision loss is formulated as
\begin{align}
\notag
\mathbb{E}_{\mathbf{I}\sim\mathcal{D}}
\Bigg[
\sum_{i=1}^{T}
\frac{\Lambda}{i}
\Big(
\beta
\left\|
\mathbf{z}_e
-
\mathrm{sg}\big[\mathbf{z}_q{(i)}\big]
\right\|_2^2
+
\left\|
\mathrm{sg}\big[\mathbf{z}_e\big]
-
\mathbf{z}_q{(i)}
\right\|_2^2
\Big)
\Bigg].
\end{align}
Similar to the quantization loss, the first term updates the semantic encoder, while the second term updates the KBs to progressively reduce the residual error. The constant $\Lambda$ controls the overall strength of progressive supervision, and the weighting factor $1/i$ assigns larger penalties to earlier refinement stages. \textcolor{black}{By progressively minimizing the stage-wise residual error, the proposed supervision explicitly regularizes intermediate quantized representations under KB reuse, encouraging successive refinement stages to gradually reduce
the residual error.} The overall training objective is defined as:
\begin{equation}
\mathcal{L}_{\mathrm{total}}
=
\mathcal{L}_{\mathrm{rec}}
+
\mathcal{L}_{\mathrm{vq}}
+
\mathcal{L}_{\mathrm{res}}.
\end{equation}

\begin{figure*}[!t]
\centering
\subfigure[PSNR]{%
    \includegraphics[width=0.235\textwidth]
    {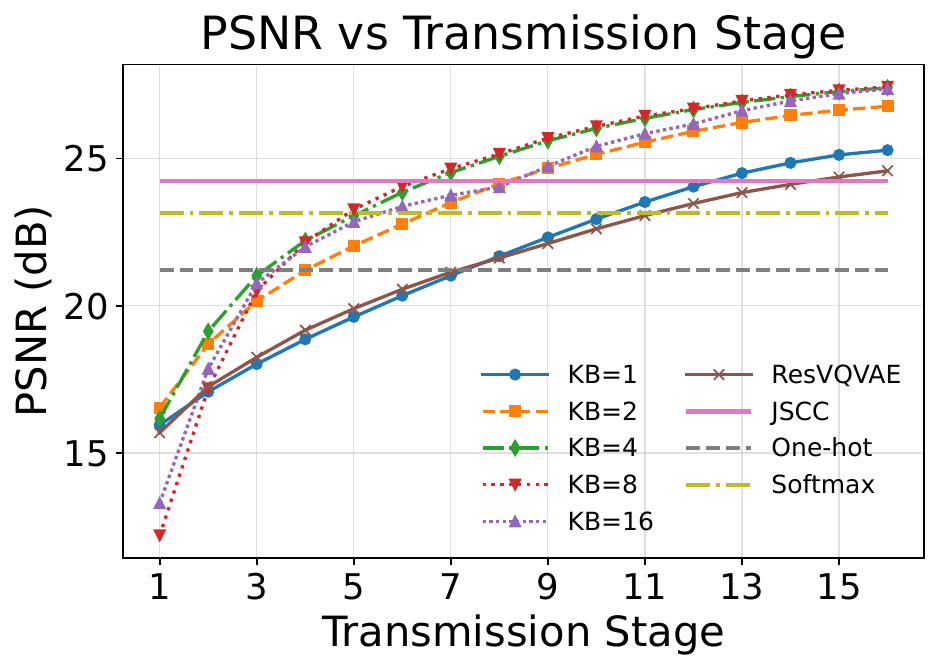}%
}\hfill%
\subfigure[SSIM]{%
    \includegraphics[width=0.235\textwidth]
    {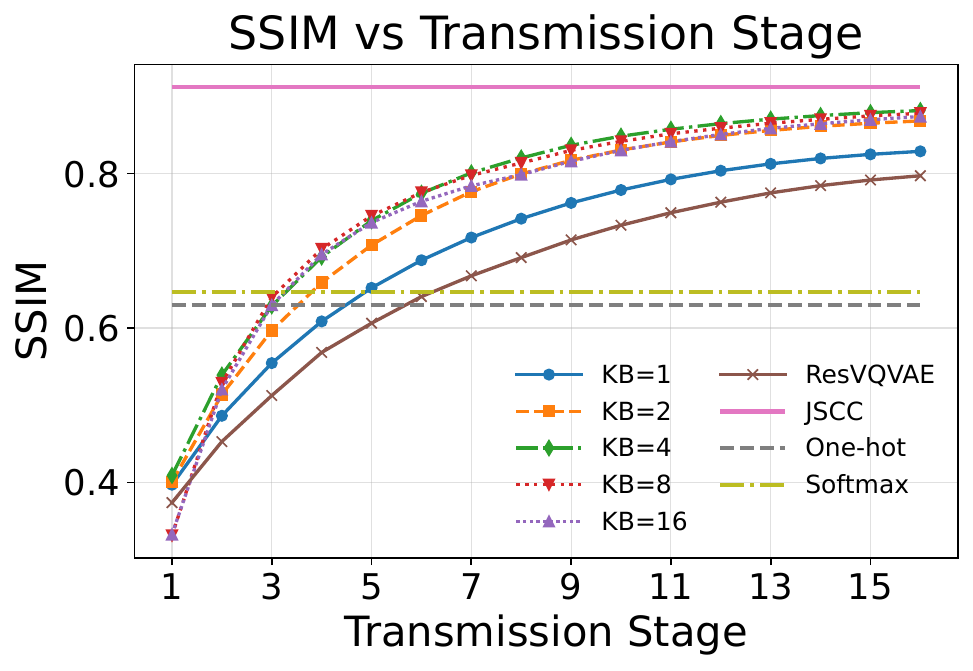}%
}\hfill%
\subfigure[KID]{%
    \includegraphics[width=0.235\textwidth]
    {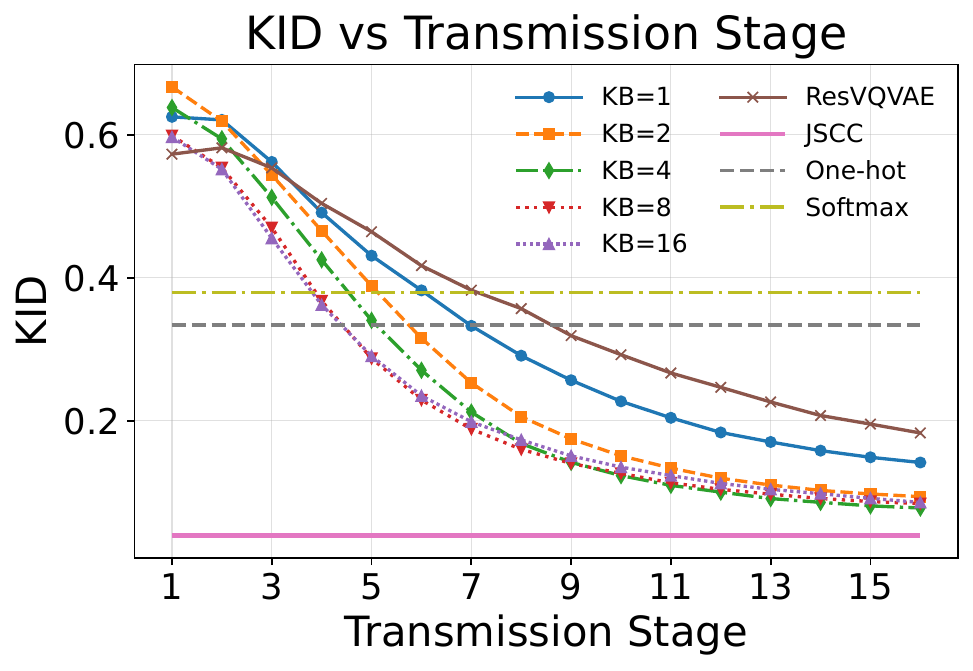}%
}\hfill%
\subfigure[FID]{%
    \includegraphics[width=0.235\textwidth]
    {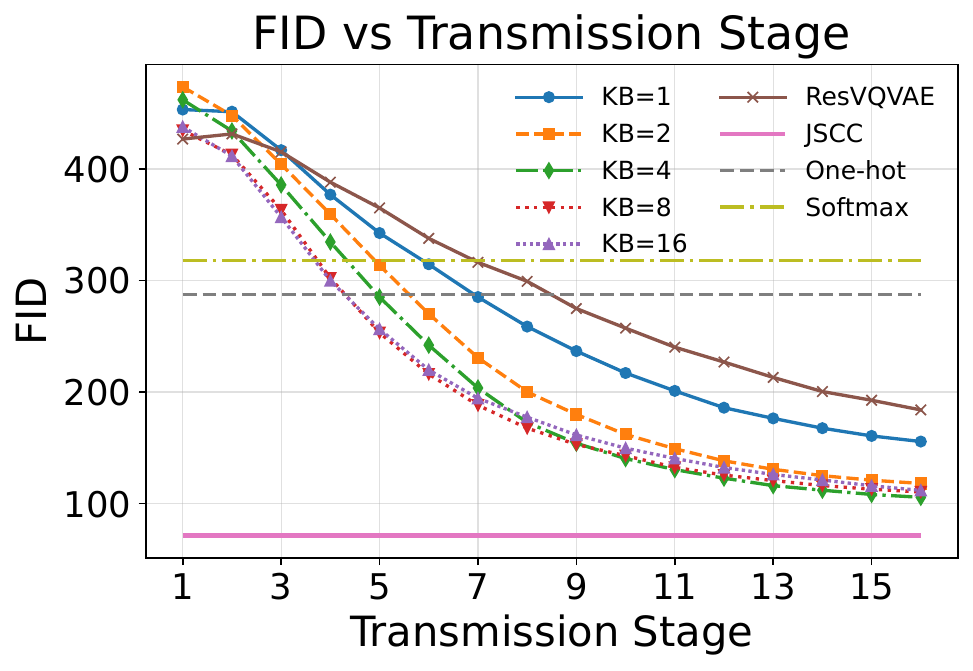}%
}
\caption{Reconstruction performance comparison on the Cityscapes dataset.}
\label{Cityscapes}
\end{figure*}

\begin{figure*}[!t]
\centering
\subfigure[PSNR]{%
    \includegraphics[width=0.235\textwidth]
    {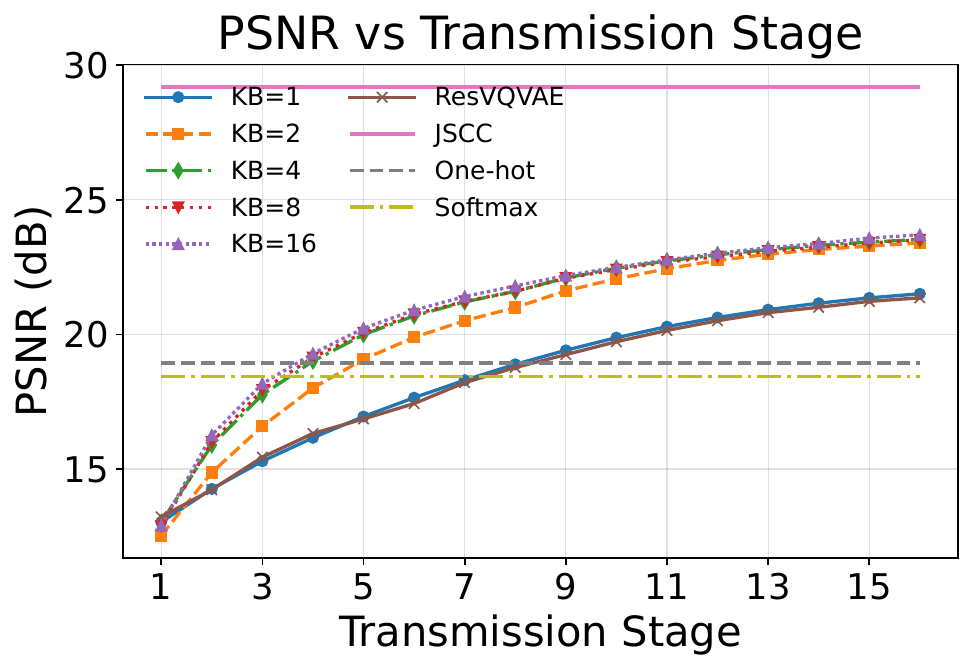}%
}\hfill%
\subfigure[SSIM]{%
    \includegraphics[width=0.235\textwidth]
    {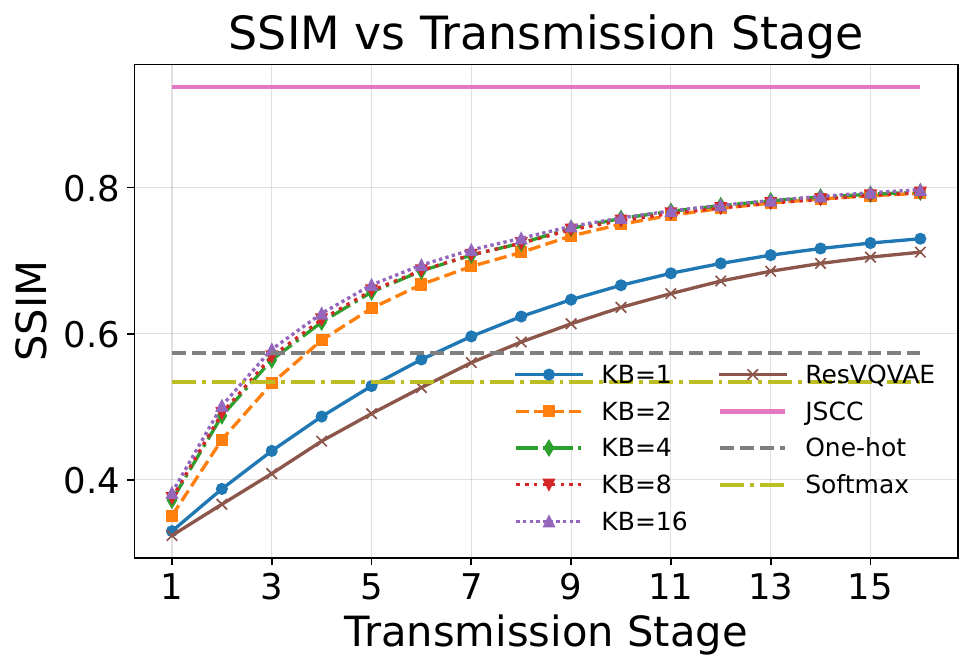}%
}\hfill%
\subfigure[KID]{%
    \includegraphics[width=0.235\textwidth]
    {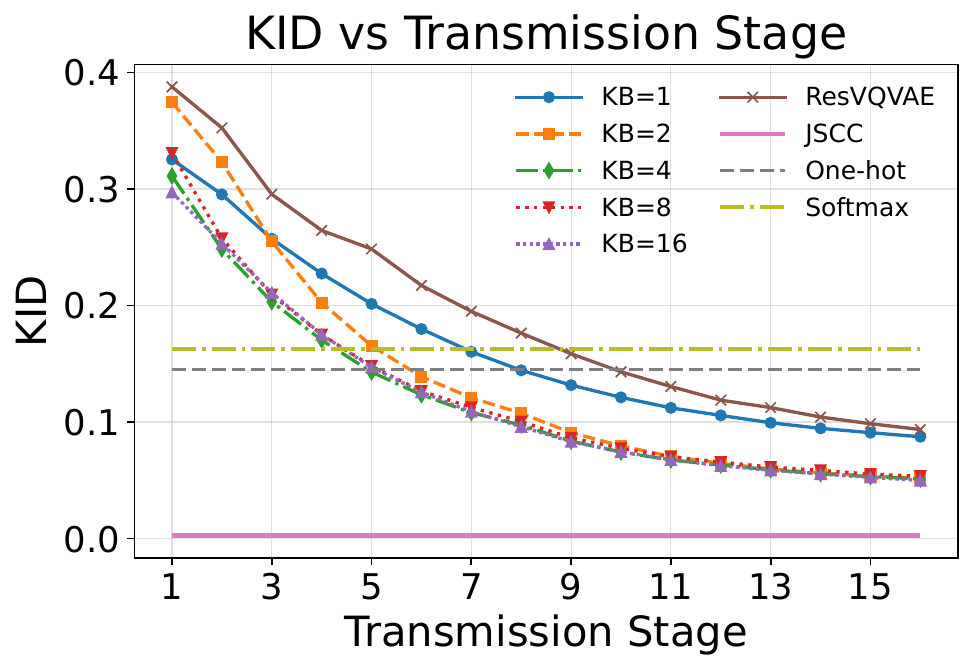}%
}\hfill%
\subfigure[FID]{%
    \includegraphics[width=0.235\textwidth]
    {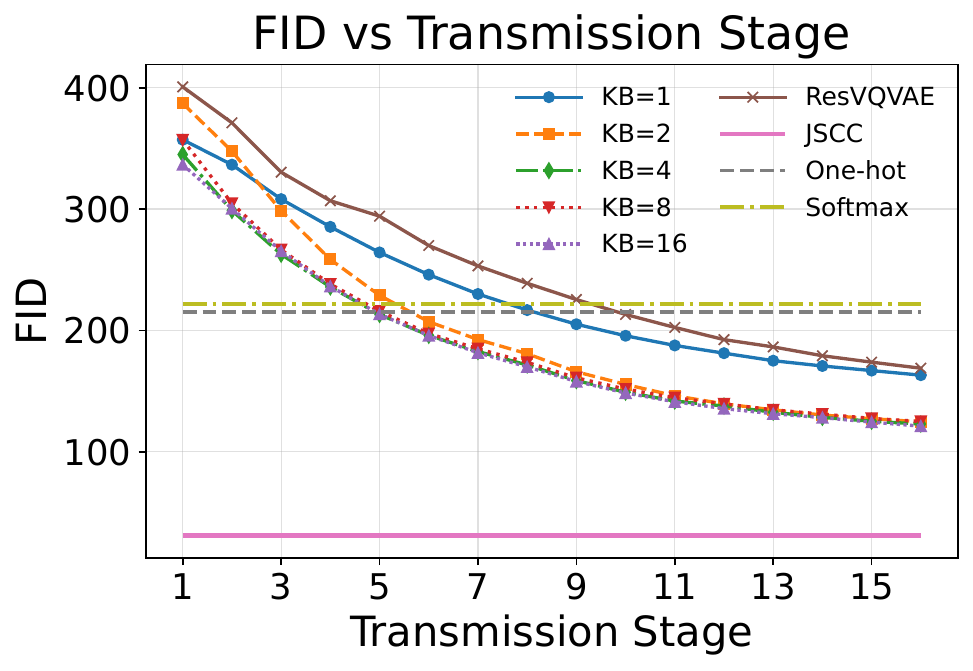}%
}
\caption{Reconstruction performance comparison on the COCO dataset.}
\label{coco}
\end{figure*}

\begin{figure*}[!t]
\centering
\subfigcapskip=0pt

\subfigure[Ground Truth]{%
\begin{minipage}[t]{0.20\textwidth}
\centering
\includegraphics[width=\linewidth]{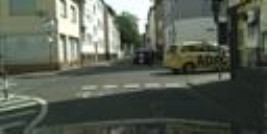}
\end{minipage}}%
\hfill
\subfigure[Stage1(SSKBQ)]{%
\begin{minipage}[t]{0.20\textwidth}
\centering
\includegraphics[width=\linewidth]{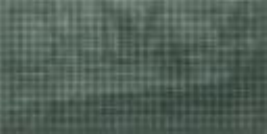}
\end{minipage}}%
\hfill
\subfigure[Stage4(SSKBQ)]{%
\begin{minipage}[t]{0.20\textwidth}
\centering
\includegraphics[width=\linewidth]{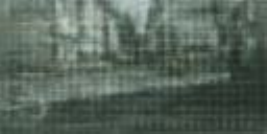}
\end{minipage}}%
\hfill
\subfigure[Stage8(SSKBQ)]{%
\begin{minipage}[t]{0.20\textwidth}
\centering
\includegraphics[width=\linewidth]{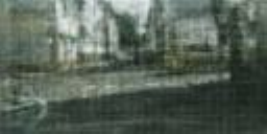}
\end{minipage}}%
\hfill
\subfigure[Stage16(SSKBQ)]{%
\begin{minipage}[t]{0.20\textwidth}
\centering
\includegraphics[width=\linewidth]{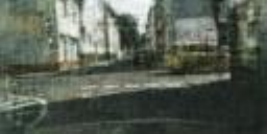}
\end{minipage}}%

\vspace{-6pt}

\subfigure[JSCC]{%
\begin{minipage}[t]{0.20\textwidth}
\centering
\includegraphics[width=\linewidth]{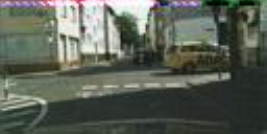}
\end{minipage}}%
\hfill
\subfigure[VQVAE(Gumbel-Softmax)]{%
\begin{minipage}[t]{0.20\textwidth}
\centering
\includegraphics[width=\linewidth]{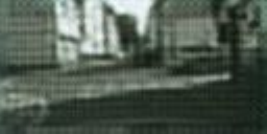}
\end{minipage}}%
\hfill
\subfigure[VQVAE(One-Hot) ]{%
\begin{minipage}[t]{0.20\textwidth}
\centering
\includegraphics[width=\linewidth]{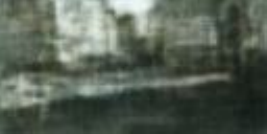}
\end{minipage}}%
\hfill
\subfigure[Stage8(MKBQ)]{%
\begin{minipage}[t]{0.20\textwidth}
\centering
\includegraphics[width=\linewidth]{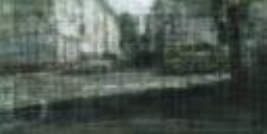}
\end{minipage}}%
\hfill
\subfigure[Stage16(MKBQ)]{%
\begin{minipage}[t]{0.20\textwidth}
\centering
\includegraphics[width=\linewidth]{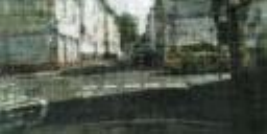}
\end{minipage}}%

\caption{Image reconstruction performance under different baselines.}
\label{sem}
\end{figure*}

\section{Simulation Results}
\subsection{Experiment Settings}

\textit{1) Datasets:}
\textcolor{black}{The Cityscapes and COCO datasets are used to evaluate the proposed SSKBQ
scheme. Cityscapes contains 2,975 training images and 500 validation images of urban street scenes, whereas COCO contains more diverse objects and visual scenes. For a consistent evaluation, images from both datasets are resized to \(128\times64\).}

\textit{2) Evaluation Metrics:}
Reconstruction quality is evaluated using PSNR~\cite{jahne2005digital}, SSIM~\cite{wang2004image}, FID~\cite{NIPS2017_8a1d6947}, and KID~\cite{bińkowski2021demystifying}.

\textit{3) Baselines:}
We compare SSKBQ with representative non-progressive and progressive
schemes. JSCC~\cite{bourtsoulatze2019deep} directly transmits continuous
semantic features and serves as an unquantized reference. For SKBQ, the
VQVAE~\cite{van2017neural} system is implemented using one-hot and
Gumbel-softmax quantization. MKBQ employs an independent KB at each
transmission stage.

\textit{4) Parameter Settings:}
\textcolor{black}{For both datasets, the input image and semantic feature
dimensions are \(3\times128\times64\) and \(256\times32\times16\),
respectively. Each KB contains \(N=512\) codewords of dimension \(c=256\),
and \(hw=512\) spatial feature vectors are quantized at each stage. We
consider 16-stage progressive transmission and set
\(\alpha=\beta=0.25\) and \(\Lambda=8\). During testing, only the codeword
indices associated with the quantized feature vectors are transmitted.
The receiver retrieves the corresponding codewords from the shared KBs
according to the received indices and reconstructs the quantized semantic
feature.}

\subsection{\textcolor{black}{Complexity and Scalability}}

\textcolor{black}{SKBQ, MKBQ, and SSKBQ require
\(O(Nc)\), \(O(TNc)\), and \(O(KNc)\) KB storage, respectively.
By reusing \(K\) KBs across \(T\) stages, SSKBQ requires only \(K/T\)
of the KB storage of MKBQ.}

\textcolor{black}{At each stage, nearest-neighbor assignment compares
\(hw\) feature vectors of dimension \(c\) with \(N\) codewords, resulting
in \(O(chwN)\) transmitter-side complexity. The cumulative complexity
after \(t\) stages is therefore \(O(tchwN)\) for both MKBQ and SSKBQ.}

\textcolor{black}{At the receiver, the transmitted indices directly identify
the codewords. Index lookup, feature assembly, and residual accumulation
require \(O(chw)\) operations per stage and \(O(tchw)\) after \(t\) stages,
independent of \(N\).}

\textcolor{black}{Let \(C_{\mathrm{dec}}\) denote the cost of one decoder
execution. Reconstruction after all \(t\) stages requires
\(O(tchw+C_{\mathrm{dec}})\), whereas reconstruction after every stage
requires \(O(tchw+tC_{\mathrm{dec}})\). Hence, progressive inference
latency is mainly determined by repeated decoder executions, while KB
processing grows only linearly with \(t\). Actual wall-clock latency also
depends on the hardware platform and implementation.}

\subsection{\textcolor{black}{Comparison with Existing Schemes}}

\textcolor{black}{
To compare the proposed SSKBQ with existing semantic communication
schemes, experiments are conducted on the Cityscapes and COCO datasets.
The progressive reconstruction results are presented in
Figs.~\ref{Cityscapes} and~\ref{coco}, where different numbers of reusable
KBs are evaluated over 16 transmission stages.
}

\textcolor{black}{
Across both datasets, SSKBQ exhibits consistent progressive reconstruction
trends. As more transmission stages are received, the reconstruction
quality gradually improves in terms of PSNR, SSIM, FID, and KID, validating
the effectiveness of KB reuse for progressive semantic refinement.
Moreover, compared with MKBQ under comparable settings, SSKBQ achieves
better reconstruction performance, demonstrating that the proposed
stage-aware residual supervision facilitates the optimization of reused
KBs across multiple refinement stages.
}

\textcolor{black}{
Compared with single-stage quantization schemes, including VQVAE(OH) and
VQVAE(GS), SSKBQ achieves consistent improvements on both datasets,
demonstrating the benefit of residual refinement. The comparison with JSCC
shows dataset-dependent behavior. On Cityscapes, SSKBQ achieves higher
PSNR after sufficient progressive stages, whereas JSCC maintains better
performance on COCO. This difference results from the interaction among
dataset complexity, semantic representation capability, and codec capacity.
JSCC avoids quantization distortion by transmitting continuous semantic
features, which is advantageous for complex image distributions. In
contrast, SSKBQ provides a more efficient progressive transmission
mechanism when semantic information can be effectively organized through
reusable KBs.
}

The influence of KB number is consistent with the observations in the
previous subsection. In early transmission stages, smaller KB
configurations provide better performance because semantic information is
more concentrated within each reusable KB. With increasing transmission
stages, larger KB configurations gradually benefit from their higher
representation capacity and achieve better final reconstruction quality.
This result highlights the necessity of balancing representation capacity
and KB scalability, which motivates the proposed KB reuse strategy.
A representative reconstruction example is shown in Fig.~\ref{sem}.

For quantitative comparison, Tables~\ref{tab1} and
\ref{tab:coco_comparison} summarize the final reconstruction performance
of different schemes on the Cityscapes and COCO datasets, respectively.
The JSCC results are highlighted in bold as the continuous-feature
transmission baseline. VQVAE(GS) and VQVAE(OH) denote the
Gumbel--Softmax and one-hot quantization implementations, respectively.
In MKBQ($t$) and SSKBQ($t$), $t$ represents the number of received
progressive transmission stages.

\textcolor{black}{To evaluate communication efficiency, the number of transmitted bits for
one image sample is adopted as the transmission rate. Specifically, JSCC directly transmits the continuous
semantic feature
$\mathbf{z}_{e}\in\mathbb{R}^{256\times32\times16}$, resulting in
8,388,608 transmitted bits under 64-bit floating-point representation.
For MKBQ and SSKBQ, only the indices of selected codewords are transmitted.
Since each KB contains $N=512$ codewords, each index requires 9 bits.
Therefore, the transmission rates for 1, 8, and 16 progressive stages are
4,608, 36,864, and 73,728 bits, respectively. For VQVAE(GS), transmitting
the 512-dimensional soft assignment weights for each spatial feature vector
requires 16,777,216 transmitted bits. These results demonstrate that
SSKBQ enables discrete-index semantic transmission with substantially
reduced communication overhead compared with continuous-feature and
soft-assignment-based schemes.}

\textcolor{black}{As shown in Table~\ref{tab1}, SSKBQ(16) achieves a PSNR of 27.36 dB on the Cityscapes dataset, improving by 11.31\%, 28.93\%, and 18.13\% over MKBQ(16), VQVAE(OH), and VQVAE(GS), respectively. Meanwhile, SSKBQ(16) requires only 73,728 transmitted bits, corresponding to less than 1\% of the transmission overhead of JSCC. Despite this substantial reduction in communication cost, SSKBQ achieves a 12.87\% PSNR improvement over JSCC. However, JSCC obtains better SSIM, FID, and KID values, which can be attributed to its direct transmission of continuous semantic features and the resulting avoidance of quantization distortion.}

\textcolor{black}{The results on the COCO dataset exhibit a different trend. As shown in Table~\ref{tab:coco_comparison}, SSKBQ(16) achieves a PSNR of 23.53 dB, outperforming MKBQ(16), VQVAE(OH), and VQVAE(GS) by 10.16\%, 24.17\%, and 27.60\%, respectively. However, JSCC still achieves higher reconstruction quality on COCO due to its stronger representation
capability for diverse object categories and complex visual distributions.
This observation indicates that SSKBQ is not intended to universally
replace continuous-feature JSCC, but rather provides a storage-efficient
and progressive semantic transmission framework that achieves a favorable
trade-off between communication overhead and reconstruction performance.}

\begin{table}[!t]
\centering
\footnotesize
\caption{Image Reconstruction Performance Comparison (Cityscapes).}
\label{tab1}
\setlength{\tabcolsep}{10pt}
\renewcommand{\arraystretch}{1.15}

\begin{tabular}{lcccc}
\hline\hline
\textbf{Method}
& \textbf{PSNR (dB)}
& \textbf{SSIM}
& \textbf{FID}
& \textbf{KID}
\\
\hline

\textbf{JSCC}
& \textbf{24.24}
& \textbf{0.9123}
& \textbf{71.38}
& \textbf{0.0398}
\\

MKBQ(1)
& 15.69
& 0.3739
& 426.77
& 0.5728
\\

MKBQ(8)
& 21.62
& 0.6912
& 299.22
& 0.3569
\\

MKBQ(16)
& 24.58
& 0.7972
& 183.94
& 0.1831
\\

SSKBQ(1)
& 25.28
& 0.8288
& 155.60
& 0.1418
\\

SSKBQ(8)
& 25.15
& 0.8137
& 167.59
& 0.1602
\\

SSKBQ(16)
& 27.36
& 0.8739
& 111.77
& 0.0860
\\

VQVAE(GS)
& 23.16
& 0.6463
& 317.84
& 0.3793
\\

VQVAE(OH)
& 21.22
& 0.6297
& 287.42
& 0.3336
\\

\hline\hline
\end{tabular}
\end{table}

\section{Conclusion}

This letter investigates the storage-performance trade-off in progressive
semantic communication with knowledge-base quantization. Conventional
single knowledge-base schemes achieve low storage overhead but suffer from
limited refinement capability, while multi-knowledge-base residual schemes
support progressive transmission at the cost of storage complexity that
scales with the number of refinement stages. To address this issue, we
propose a storage-scalable knowledge-base reuse quantization scheme that
decouples the number of maintained knowledge bases from the number of
progressive transmission stages. By reusing a limited number of knowledge
bases across multiple refinement stages, SSKBQ reduces storage requirements
while preserving progressive reconstruction capability.

Experimental results on different image datasets demonstrate that SSKBQ
achieves competitive reconstruction performance with substantially reduced
storage overhead compared with existing knowledge-base quantization
schemes. Moreover, the results reveal that increasing the number of
knowledge bases does not always guarantee better performance, since
semantic fragmentation and optimization difficulty may limit their
effective utilization. Developing more effective training strategies for
multi-KB architectures is therefore an important direction for future
research. Furthermore, the current study focuses on semantic-layer KB
reuse under reliable index delivery. \textcolor{black}{Extending SSKBQ to practical communication environments, including noisy index transmission, fading, packet loss, and channel-aware KB adaptation, will be investigated in future work.}

\begin{table}[!t]
\centering
\footnotesize
\caption{Image Reconstruction Performance Comparison (COCO)}
\label{tab:coco_comparison}
\setlength{\tabcolsep}{10pt}
\renewcommand{\arraystretch}{1.15}

\begin{tabular}{lcccc}
\hline\hline
\textbf{Method}
& \textbf{PSNR (dB)}
& \textbf{SSIM}
& \textbf{FID}
& \textbf{KID}
\\
\hline

\textbf{JSCC}
& \textbf{29.20}
& \textbf{0.9379}
& \textbf{30.96}
& \textbf{0.0027}
\\

MKBQ(1)
& 13.22
& 0.3238
& 400.74
& 0.3876
\\

MKBQ(8)
& 18.77
& 0.5886
& 238.82
& 0.1763
\\

MKBQ(16)
& 21.36
& 0.7116
& 168.82
& 0.0936
\\

SSKBQ(1)
& 13.02
& 0.3721
& 344.97
& 0.3111
\\

SSKBQ(8)
& 21.60
& 0.7240
& 171.49
& 0.0974
\\

SSKBQ(16)
& 23.53
& 0.7939
& 122.80
& 0.0512
\\

VQVAE(GS)
& 18.44
& 0.5341
& 221.64
& 0.1627
\\

VQVAE(OH)
& 18.95
& 0.5734
& 215.05
& 0.1450
\\

\hline\hline
\end{tabular}
\end{table}

\bibliographystyle{IEEEtran}
\bibliography{cite}

\end{document}